\documentclass[a4paper, 10pt, conference]{ieeeconf}      
\IEEEoverridecommandlockouts                              
\IEEEoverridecommandlockouts                              

\usepackage{times}

\usepackage{multicol}
\usepackage{hyperref}
\hypersetup{colorlinks,allcolors=black}

\usepackage{times}
\usepackage{graphicx} 
\usepackage{pifont}
\usepackage[ruled]{algorithm2e}  
\usepackage{caption} 
\usepackage{subcaption} 
\usepackage{tabularx}
\usepackage{color} 
\usepackage{amsmath} 
\usepackage{amssymb}  
\usepackage{tikz-cd}
\usepackage{multirow}
\usepackage{multicol}
\let\labelindent\relax
\usepackage{enumitem}
\usepackage[bottom]{footmisc} 
\usepackage{hyperref}

\usepackage{stfloats} 
\usepackage{mathtools}
\usepackage{booktabs,etoolbox} 
\DeclareMathOperator*{\argmin}{\arg\!\min}
\DeclareMathOperator*{\arginf}{\arg\!\inf}

\newcommand\blfootnote[1]{%
  \begingroup
  \renewcommand\thefootnote{}\footnote{#1}%
  \addtocounter{footnote}{-1}%
  \endgroup
}

\newtheorem{theorem}{Theorem}
\newtheorem{observation}{Observation}

\newtheorem{premise}{Premise}

\graphicspath{{figs/}} 

\begin{document}
\bstctlcite{IEEEexample:BSTcontrol}
\title{Online Domain Adaptation for Occupancy Mapping}

\author{Author Names Omitted for Anonymous Review. Paper-ID [add your ID here]}

\author{\authorblockN{Anthony Tompkins*}
\authorblockA{School of Computer Science\\
The University of Sydney\\
NSW, Australia.\\
anthony.tompkins@sydney.edu.au}
\and
\authorblockN{Ransalu Senanayake*}
\authorblockA{Department of Aeronautics\\ and Astronautics
\\Stanford University, CA, USA\\
ransalu@stanford.edu}
\and
\authorblockN{Fabio Ramos}
\authorblockA{School of Computer Science\\
The University of Sydney, Australia.\\
NVIDIA Research, Seattle, WA, USA.\\
fabio.ramos@sydney.edu.au}
}


%

\maketitle

\begin{abstract}
Creating accurate spatial representations that take into account uncertainty is critical for autonomous robots to safely navigate in unstructured environments. Although recent LIDAR based mapping techniques can produce robust occupancy maps, learning the parameters of such models demand considerable computational time, discouraging them from being used in real-time and large-scale applications such as autonomous driving. Recognizing the fact that real-world structures exhibit similar geometric features across a variety of urban environments, in this paper, we argue that it is redundant to learn all geometry dependent parameters from scratch. Instead, we propose a theoretical framework building upon the theory of \emph{optimal transport} to adapt model parameters to account for changes in the environment, significantly amortizing the training cost. Further, with the use of high-fidelity driving simulators and real-world datasets, we demonstrate how parameters of 2D and 3D occupancy maps can be automatically adapted to accord with local spatial changes. We validate various domain adaptation paradigms through a series of experiments, ranging from inter-domain feature transfer to simulation-to-real-world feature transfer. Experiments verified the possibility of estimating parameters with a negligible computational and memory cost, enabling large-scale probabilistic mapping in urban environments.
\blfootnote{
\noindent *Equal contribution \\
\noindent Video: \href{https://youtu.be/qLv0mM9Le8E}{https://youtu.be/qLv0mM9Le8E} \\
\noindent Code: \href{https://github.com/MushroomHunting/RSS2020-online-domain-adaptation-pot}{github.com/MushroomHunting/RSS2020-online-domain-adaptation-pot} \\
\noindent Appendix: \href{https://github.com/MushroomHunting/RSS2020-online-domain-adaptation-pot}{github repository}
}
\end{abstract}

\IEEEpeerreviewmaketitle

\section{Introduction}

\begin{figure*}[b]
    \centering  
    \includegraphics[width=1.0\linewidth]{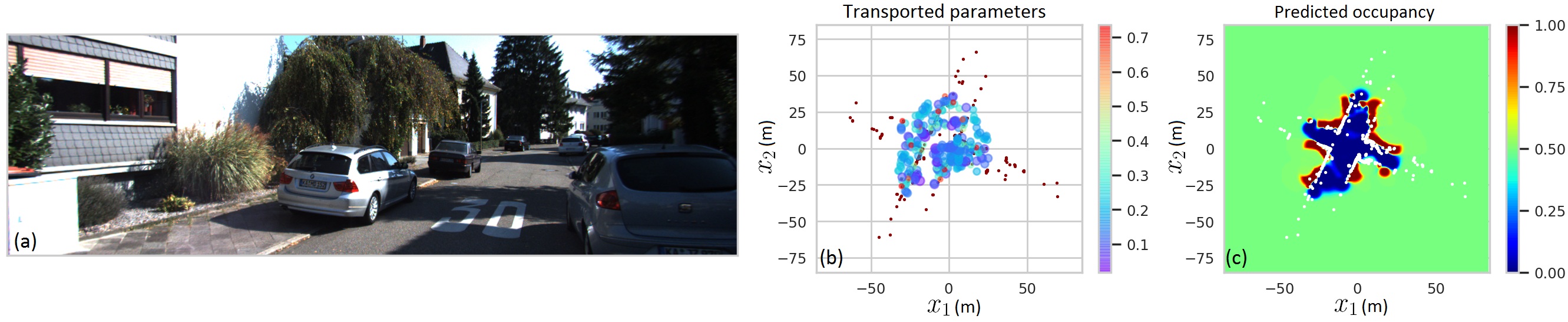}
    \caption{ (a) Forward camera-view from a car that has just passed an urban intersection (KITTI dataset). (b) A set of occupancy model parameters estimated using the proposed Parameter Optimal Transport (POT) method. Values of these parameters depend on the geometry of the environment. Note that these parameters were \textit{transferred online} from a simulated environment and were \textit{never learned from scratch}. (c) Mean occupancy map obtained from the transferred parameters.}
    \label{fig:kitti_frame}
\end{figure*}

The demand for intelligent robots in day-to-day activities is growing as never before. However, one of the main reasons hindering the deployment of robots in real-world environments is the challenge of reliably adapting to continuously changing environments. Since a robot typically represents its environment and itself using mathematical models, it is indispensable to adjust these models to accommodate changes in the environment the robot operates in. For instance, if the model is represented as a parameterized statistical model, its parameters should be regularly redetermined to adjust for changes to new environments and data. 

If the learning procedure is computationally expensive, frequently updating the model parameters in real-time is a significant challenge. This is indeed the case in deep learning as well as in many Bayesian inference techniques. While there are many methods to adapt deep neural networks to varying domains \cite{ammar2015autonomous,sadeghi2017sim2real,finn2017model,fang2018dart,wulfmeier2018incremental}, such adaptation techniques are under-explored for Bayesian models \cite{meier2014efficient} despite their extensive applications in robotics \cite{deisenroth2011pilco,wuthrich-rgf-2015,campbell2017bayesian,burchfiel2017bayesian,unhelkar2018learning}. As uncertainty is represented as probability distributions in Bayesian models, entire distributions need to be adapted when changing to a new domain. The question remains: how do we solve the problem of efficient adaptation without retraining models from scratch? In this paper, we focus on learning the uncertainty of occupancy in an unknown environment by transferring model parameters associated with a source dataset to a target dataset in a zero-shot fashion \cite{isele2016using}. This transfer procedure significantly reduces the time to estimate the model parameters, as opposed to learning them from scratch.

Even though the fundamental techniques developed in this paper have great potential to be used in a variety of data-efficient robot perception and planning applications, our focus is to build an online continuous mapping method for arbitrarily large environments. Our formulation builds upon the state-of-the-art Bayesian occupancy mapping technique named automorphing Bayesian Hilbert maps (ABHMs) \cite{tomkins2018nonstationar}. By developing a novel parameter transfer learning technique, we make this theoretically rich, yet practically less scalable offline mapping technique, run online in large-scale unknown urban environments. Since ABHM explicitly provides uncertainty estimates of which areas of the environment are occupied, it can be utilized in safety-critical robotics applications \cite{lasota2017survey} such as autonomous driving. For instance, they can be integrated into safe-motion planning algorithms and risk-aware decision-making in cluttered and dynamic real-world urban environments \cite{akametalu2014reachability,vallicrosa2018h}.
The main reason that hinders the use of ABHM in real-world applications is the run-time cost of learning parameters as it relies on an expensive black-box variational inference technique. Because these parameters are spatially local and depend on the geometrical features of the objects in the environment, parameters in one location of the environment are completely different from another. Therefore, ABHM requires learning these spatially variant parameters for every location of the environment. Moreover, in dynamic environments, these parameters need to be swiftly adjusted to the changing occupancy level. Taking into account these limitations, it is essential to quickly estimate the parameters in an alternative and more efficient manner.

As an alternative to relearning parameters in a new scene, we propose to transfer ``geometry-dependent spatial features'' of the ABHM model from a training data pool to the current scene. We show that this can be efficiently done using the theory of Optimal Transport \cite{villani2008optimal}, which recently regained popularity due to its successful application to several machine learning algorithms \cite{arjovsky2017wasserstein,solomon2015convolutional}. The proposed approach completely bypasses explicitly learning parameters of the statistical model which are typically learned through a complicated log-likelihood loss. In essence, as shown in Figure~\ref{fig:kitti_frame}, the algorithm ``transports'' location and geometry-dependent parameters of the model from one place to another place by examining the similarities among LIDAR scans. This parameter transport procedure exploits geometry-dependent kernels with less computational cost, resulting in a higher quality maps. With this, we bring the following contributions,

\begin{enumerate}
    \item a theoretical framework for parameter transfer in robotics;
    \item intra-domain transfer: sequentially building a map based on features learned in previous time frames;   
    \item inter-domain transfer: mapping an environment with features learned from another environment. This includes parameter transfer from one town to another, static to dynamic environments, and simulation to real-world; and
    \item online and efficient mapping of large-scale 2D and 3D environments.
\end{enumerate}

Notation given in Table~\ref{table:notation} will be used throughout the paper. 
\begin{table}[h]
  \centering
    \caption{Table of notations and terminology}
    \setlength{\tabcolsep}{2pt} 
    \begin{tabular}{c|l}
    \toprule
       Notation & \multicolumn{1}{|c}{Description} \\
      \hline    
      \rule{0pt}{3ex}
      $\bar{}$ and $\breve{}$ & Mean and variance of Gaussian; shape and scale of Gamma\\
      $\mathbf{x}$ and $y$ & LIDAR data positions and labels  \\
    $N$ and $M$ & Number of data points and number of parameters \\
    $\mathbf{\bar{h}}$ & Kernel positions\\
    $\theta$ & Parameter set except $\mathbf{\bar{h}}$\\
      $^{(\mathcal{S})}$ and $^{(\mathcal{T})}$ & Source and target \\
      $P$ & Coupling matrix\\
      $a \to b$ & transport = transfer = domain adaptation = transform \\ & \quad = map = convert (from a to b) \\
    \bottomrule
    \end{tabular}
  \label{table:notation}
\end{table}

\section{Preliminaries}
\label{sec:prelim}

\subsection{Uncertainty of Occupancy}
\label{sec:bhm}

\begin{figure}[]
    \centering
    \includegraphics[width=0.8\linewidth]{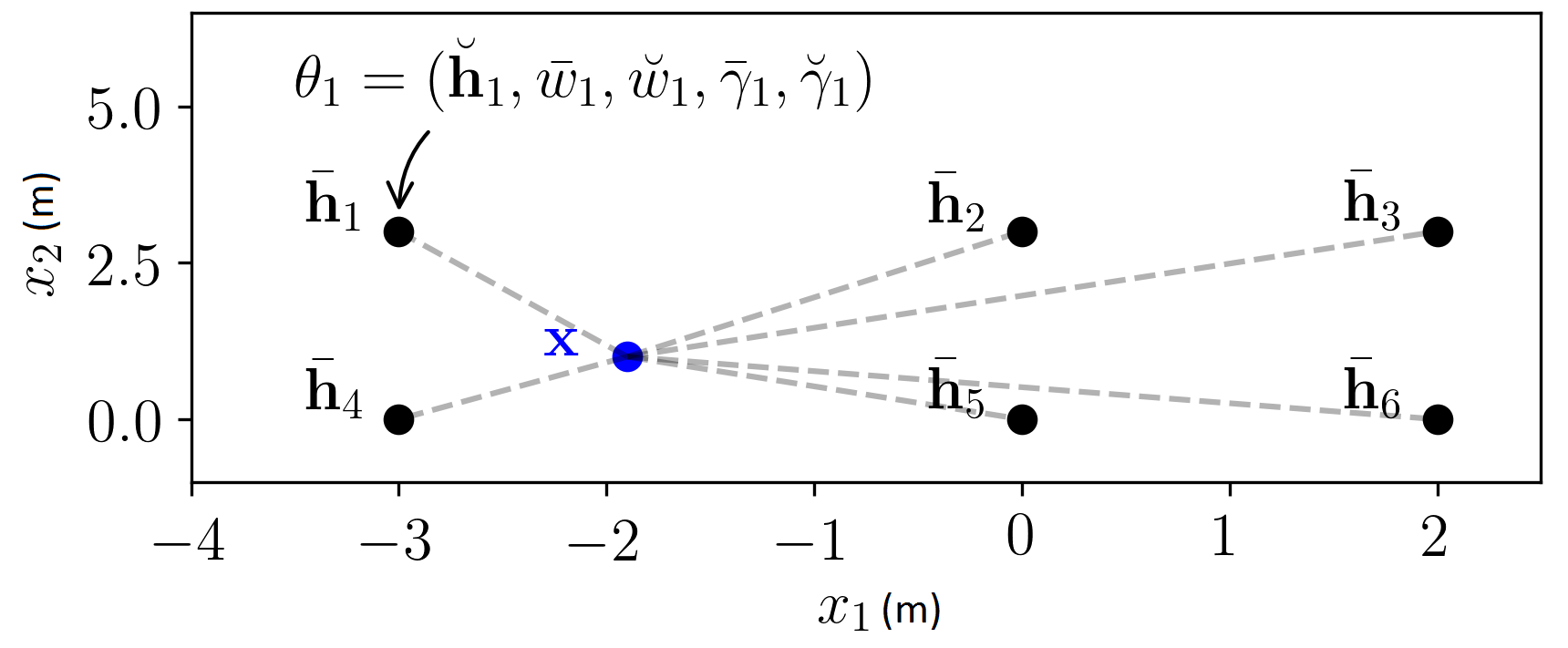}
    \caption{Kernel positioning. Kernels are placed in different locations $\mathbf{\bar{h}}$. For instance, here, the distance between each data point $\mathbf{x}$ and  $\{\mathbf{\bar{h}}_m \}_{m=1}^{M=6}$ has to be evaluated as in eq.~\ref{eq:bhm}. }
    \label{fig:kernel_exp}
\end{figure}

\begin{figure*}[]
    \centering
    \includegraphics[width=1.0\linewidth]{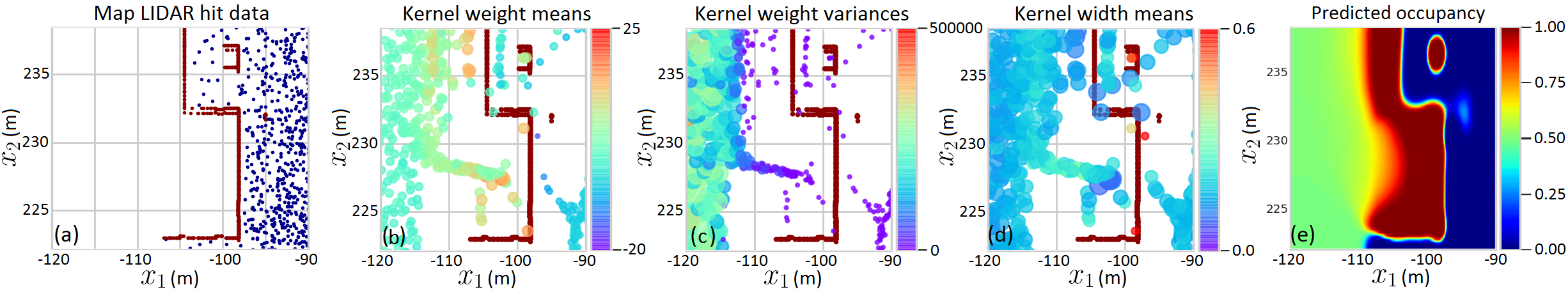}
    \caption{Spatial correlation among obstacles in the environment and some ABHM parameters. (a) LIDAR data: $y=1$ (hits) in red and $y=0$ in blue. (b)-(d) kernel weight means $\bar{w}_m$, weight variances $\breve{w}_m$, and width means $\bar{\gamma}_m$. Each point is a kernel placed in the shown location $\mathbf{\bar{h}}$ in the $x_1$-$x_2$ space.  Refer Observation~\ref{obs:kernels} for further interpretation. (e) Predicted occupancy.}
    \label{fig:kernel}
\end{figure*}

An occupancy model is typically represented as a parameterized function that models the occupancy probability of each location in the environment. The objective is to learn the model parameters $\theta$ given a set of observations from LIDAR beams. Once the parameters are estimated, it is possible to query $y_* = p(\mathrm{occupied} \vert \mathbf{x}_*, \theta) \in [0,1]$ anywhere in the 2D space\footnote{We limit our discussion to 2D for simplicity. All theory are readily extensible to 3D.} $\mathbf{x}_* \in \mathbb{R}^2 := (x_1,x_2)$. Labeling LIDAR hits as $y=1=\mathrm{occupied}$ and randomly sampled points between each LIDAR hit and the LIDAR sensor as $y=0=\mathrm{free}$, a dataset $\mathcal{D}=\{ (\mathbf{x}_n,y_n)\}_{n=1}^N$ can be generated. Here, $\mathbf{x}_n \in \mathbb{R}^2$ are the corresponding spatial locations of $y_n \in \{0,1\}$.

Various models have been proposed for the occupancy function. Gaussian process occupancy maps (GPOMs) \cite{GPOMIJRR,wang2016fast} have been presented as an alternative to improve occupancy grid mapping (OGM) \cite{ElfesThesis, arbuckle2002temporal} and Hilbert maps \cite{Ramos15}. In addition to considering neighborhood information for accurate occupancy predictions, kernel methods used in GPOMs come with the flexibility of incorporating other aspects such as dynamics into occupancy mapping \cite{rans2017,senanayake2016spatio}. On the other hand, GPOMs account for uncertainty as they are based on a Bayesian nonparametric model. Regardless of their attractive theoretical properties, GPOMs are impractical for real-world usage because of the $\mathcal{O}(N^3)$ run-time and memory complexity. Recently proposed Bayesian Hilbert maps (BHMs) \cite{senanayake2017bayesian}, on the other hand, encompass all positive traits of GPOMs but at a cost of $\mathcal{O}(M^3)$ where $M \ll N$ is the number of features that correlates with the accuracy. Since ABHM considers the full Bayesian treatment over parameters of \cite{senanayake2017bayesian} to account for local spatial changes in the environment, it achieves a significantly higher accuracy.

BHM can be summarized as performing Bayesian logistic regression in a high-dimensional feature space $\mathbb{R}^M$ using {\emph kernels} \cite{hofmann2008kernel,senanayake2018building}. BHM uses the same kernel for the entire map. ABHM is an extension to BHM to learn all location-dependent nonstationary kernel parameters (Appendix~I-C). While BHM can be run in near real-time in an online fashion, ABHM is computationally expensive as it requires learning thousands of parameters offline. In ABHM, the occupancy probability of a point $\mathbf{x}_*$ is given by,
\begin{equation}
p(y_*=1 \vert \mathbf{x}_*) = \mathrm{sigmoid}\bigg( \sum_{m=1}^M w_m \underbrace{\exp{\big( -\gamma_m \| \mathbf{x}_* - \mathbf{h}_m \|_2^2 \big)} }_{m^\text{th} \text{ SE kernel}} \bigg),
\label{eq:bhm}
\end{equation}
where $w,\mathbf{h},$ and $\gamma$ are parameters learned from data $\mathcal{D}$. The inner part of the equation is a $w$ weighted sum of $M$ kernels placed in 2D spatial locations $\mathbf{h}$. In areas where there are more LIDAR hits in the locality of a kernel, then its associated weight $w_m$ will be higher, and vice versa. This is because, as illustrated in Figure~\ref{fig:kernel}, here, $M$ squared-exponential (SE) kernels positioned at mean locations $(\mathbf{\bar{h}}_1, \mathbf{\bar{h}}_2, \dots, \mathbf{\bar{h}}_M)$ are used to project 2D data into an $M$ dimensional vector such that each kernel has more effect from data in its locality. $\gamma$ are positive parameters that control the width of each kernel. Probability distributions
$w_m \sim \mathcal{N}(\bar{w}_m, \breve{w}_m)$, $\mathbf{h}_m \sim \mathcal{N}(\mathbf{\bar{h}}_m, \mathbf{\breve{h}}_m)$, and $\gamma_m \sim \mathrm{Gamma}(\bar{\gamma}_m, \breve{\gamma}_m)$ are induced on the parameters to naturally encode uncertainty. Here, slightly abusing standard notations, $\bar{}$ and $\breve{}$ symbols are used to represent the mean and dispersion parameters, respectively (Table~\ref{table:notation}). 

The parameters of the model are learned using variational inference \cite{tomkins2018nonstationar}. See Figure~\ref{fig:kernel} for some of the estimated parameters. Since there are $8$ parameters ($\bar{w}_m, \breve{w}_m, \bar{\gamma}_m, \breve{\gamma}_m \in \mathbb{R}$ and $\mathbf{\bar{h}}_m, \mathbf{\breve{h}}_m \in \mathbb{R}^2$) associated with each kernel, it is required to learn $8M$ parameters. In order to achieve a practically satisfactory accuracy to cover a 100 m$^2$ area, it is necessary to have over 10000 kernels which would take around 10 minutes on a GPU. On the other hand, although ABHM provides high-quality maps, it is required to first collect the entire dataset as it does not support sequential training, making it practically unsuitable for mobile robotics applications.

\subsection{Domain Adaptation}
\label{sec:da}

The learned model parameters for a sample environment can be visualized in Figure~\ref{fig:kernel}. 

\begin{observation} 
\label{obs:kernels}
Once the full ABHM model is learned, the following can be observed: 
\begin{enumerate}[topsep=0pt,itemsep=-1ex,partopsep=1ex,parsep=1ex]
    \item As shown in Figure~\ref{fig:kernel} (b), the mean values of weights $\bar{w}$ are higher in areas where there are LIDAR hits, and vice versa. In areas where there are no observations at all ($x_1 \lessapprox -105$ in Figure~\ref{fig:kernel} (a)), the variance values $\breve{w}$ are high as shown in Figure~\ref{fig:kernel} (c).
    \item The mean widths $\bar{\gamma}$, as can be onserved in Figure~\ref{fig:kernel} (d), are higher close to the obstacles, indicating sharp edges.
    \item The mean positions of kernels $\mathbf{\bar{h}}$ align according to the geometry of the obstacles (Figure~\ref{fig:kernel} (b)-(d)).
\end{enumerate}
\end{observation} 

\begin{premise} 
\label{arg:kernels}
Based on Observation~\ref{obs:kernels}, there is geometric correspondence between parameter values and obstacles observed by the LIDAR. Therefore, we argue that spatially dependent parameters for a new environment, defined as the target domain, can be estimated by discovering correspondence between the target (new) LIDAR data and source (known) LIDAR data with associated parameters. Here, the source is an environment whose parameters are known or pre-estimated using a method such ABHM in a simple environment, and the target is a complex and large environment whose parameters are not known and challenging to estimate. This requires transferring features from source to target domains.
\end{premise} 

Transferring knowledge obtained from one domain to the other has been widely discussed in the machine learning literature \cite{ ammar2015autonomous,pan2009survey}. The broader class of transferring from one type of domain to the other, e.g. images to text, is known as transfer learning. If the type of source and target domains are the same, as in occupancy mapping, the transfer process is called domain adaptation (DA). Applications in robotics include transferring control policies from simulation to real-world \cite{sadeghi2017sim2real, bousmalis2018using}, and making image processing tasks invariant to lighting and other changes \cite{wulfmeier2018incremental,hoffman2016fcns}.

Variations of generative adversarial networks (GANs) such as DTN \cite{ghifary2016deep}, CycleGAN \cite{zhu2017unpaired}, DiscoGAN \cite{kim2017learning}, UNIT \cite{liu2017unsupervised}, DART \cite{fang2018dart} have been widely used for domain adaptation of RGB images. However, not only do these methods require a large amount of data but also it is not immediately clear how to use these techniques with sparse LIDAR data nor transferring probability distributions. In the next section, we consider an alternative domain adaptation method based on optimal transport (OT) \cite{villani2008optimal} to transfer parameters of the Bayesian occupancy model using sparse LIDAR data.

\section{Optimal Parameter Transport}
\label{sec:OT}

In this section, we present the proposed algorithm. Transferring parameters is a two-step procedure: creating a source dataset offline (Section~\ref{sec:prep}) and transferring them to a target domain online (Section~\ref{sec:pot}). Section~\ref{sec:tda} is a generalization and is the actual algorithm used in experiments. Section~\ref{sec:repot} is an extension to further improve the map quality.

\subsection{Preparing the Source Dictionary of Atoms}
\label{sec:prep}

\begin{figure}[h]
    \centering
    \includegraphics[width=1.0\linewidth]{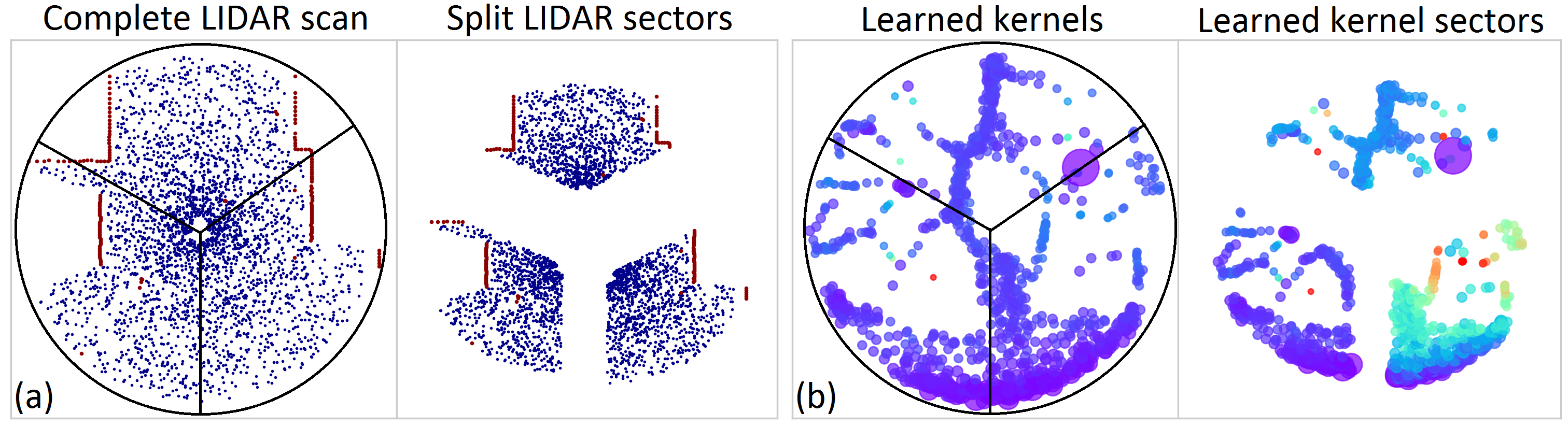}
    \caption{Extracting source data. (a) Splitting source LIDAR scans into 3 sectors. (b) Corresponding kernels parameters are also split the same way. Only kernel position means and weight means are shown here.}
    \label{fig:train_chunks}
\end{figure}

In order to take advantage of domain adaptation we must have accurately pre-trained maps from which we can extract spatially relevant features. In the context of our problem, we must extract LIDAR scans (hits and free) with their corresponding model parameters including kernel weights, positions, and widths. To provide high-quality training data we extract learned model parameters from ABHM maps. Since ABHM can only be used on small areas due to the high computational cost, we learn separate ABHM maps for different areas and construct a dictionary of source atoms which we call a {\em dictionary of atoms}.

To construct the dictionary, as illustrated in Figure~\ref{fig:train_chunks}, we split each LIDAR scan into circular sectors with radii equal to the specified maximum LIDAR distance. Rather than using the entire LIDAR scan as the source dataset, this split not only results in a diverse set of geometric primitives but also provides simpler sources for the transfer procedure presented in the following section. The corresponding learned model parameters for each sector are considered as source parameters that we wish to transfer to the target domain. For each sector, we have $M^{(\mathcal{S})}$ parameters $\{ \theta_m^{(\mathcal{S})} \}_{m=1}^{M^{(\mathcal{S})}}$ associated with ${N^{(\mathcal{S})}}$ LIDAR hits or free points $\{ (\mathbf{x}_n^{(\mathcal{S})}, y_n^{(\mathcal{S})}) \}_{n=1}^{N^{(\mathcal{S})}}$. The collection of these different LIDAR sectors constitutes the dictionary of source atoms $\mathcal{X}^{(\mathcal{S})}$. 


\subsection{Source to Target Parameter Transport}
\label{sec:pot}

Until we present the general transfer procedure that we used in experiments in Section~\ref{sec:tda}, for the sake of simplicity of the following discussion, let us assume that the dictionary of atoms contains only one LIDAR sector and associated parameters.

\begin{figure*}[b]
    \centering
    \includegraphics[width=1.0\linewidth]{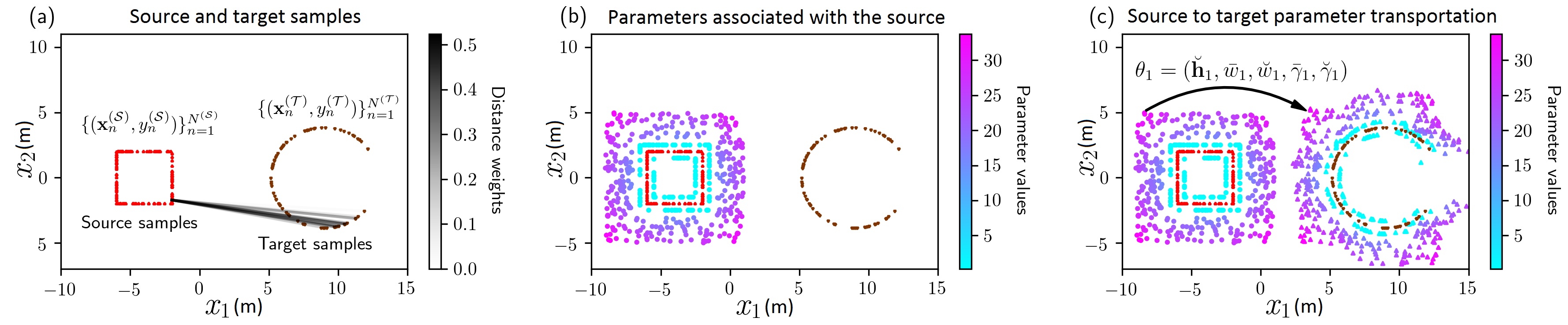}
    \caption{Optimal transport from a square to an arc. (a) If there are $N^{(\mathcal{S})}$ and $N^{(\mathcal{T})}$ number of data points in the source (red) and target (brown) datasets, the coupling matrix $\gamma$ is size $N^{(\mathcal{S})} \times N^{(\mathcal{T})}$ where any column or any row sums to $1$. A given row in $\gamma$ indicates the probabilities of the sample associated with that row could be coupled to all samples in the target dataset. Probabilities associated with one such source point to target matches are shown in white-black color scale. Note that only the 10 highest matches are shown for clarity. (b) For a given set of LIDAR hits (red) spatial parameters can be learned using ABHM. Here we see kernel parameters spread across the environment. However, for another set of LIDAR hits (brown) we would prefer not re-learning parameters because it is expensive. (c) Based on the coupling matrix between the source and the target, we transport (move from the target area to the source area) the parameters around each point. Note that how the small lengthscales (cyan) stays close to the LIDAR hits and larger lengthscales (magenta) move away from the LIDAR. }
    \label{fig:ot}
\end{figure*}

\begin{figure}[]
    \centering
    \includegraphics[width=1\linewidth]{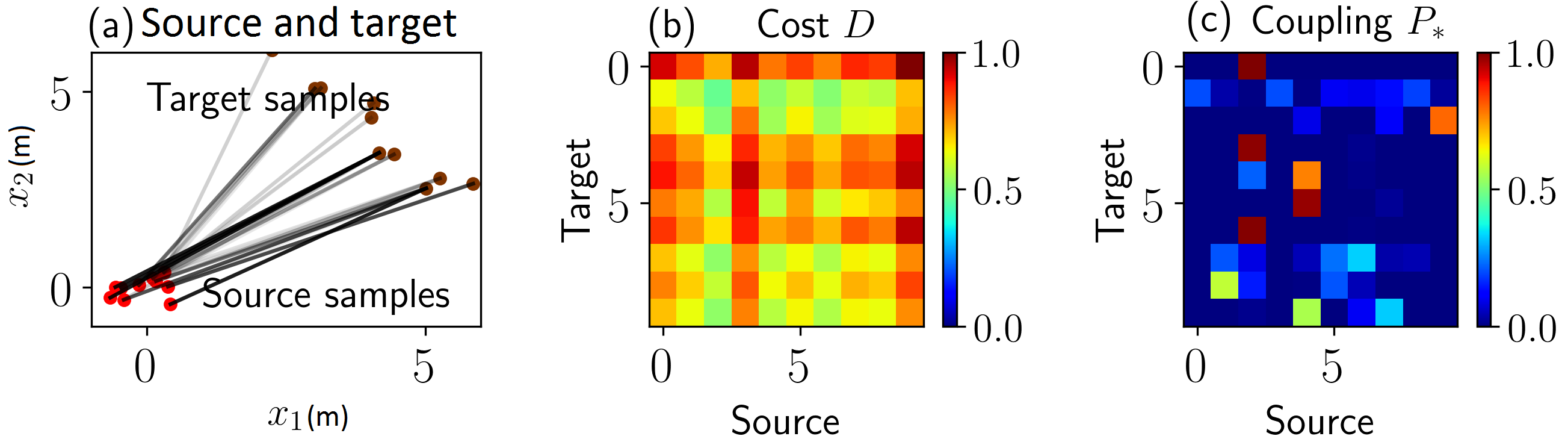}
    \caption{(a) 10 red and 10 brown dots indicate samples in $\mathbb{R}^2$ from the bivariate source and target distributions, respectively. The higher the transparency of gray lines, the lower the probability of couplings (matches) obtained after solving eq.~\ref{eq:sink}. (b) $10 \times 10$ pairwise cost matrix $D$ between the positions of samples. (c) $10 \times 10$ coupling matrix $P_*$ indicates the optimal coupling probability of source points and all other target points. Determining this matrix (and gray lines in (a)) is the goal of optimal transport.}
    \label{fig:costmap}
\end{figure}

{\bf Objective}: Having determined source LIDAR data $\{ (\mathbf{x}_n^{(\mathcal{S})}, y_n^{(\mathcal{S})}) \}_{n=1}^{N^{(\mathcal{S})}}$ and corresponding parameters $\{ \theta_m^{(\mathcal{S})} \}_{m=1}^{M^{(\mathcal{S})}}$, our objective is to  determine the new set of parameters $\{ \theta^{(\mathcal{T})} \}_{m=1}^{M^{(\mathcal{T})}}$ for a new LIDAR dataset $\{( \mathbf{x}_n^{(\mathcal{T})}, y_n^{(\mathcal{T})}) \}_{n=1}^{N^{(\mathcal{T})}}$. This problem is illustrated in Figure~\ref{fig:ot} (a) and (b). In other words, we are looking for a nonlinear mapping technique to convert a source $(\mathcal{S})$ to a target $(\mathcal{T})$. We recognize this as an \emph{optimal transport (OT) problem} given in Theorem~\ref{thm:mk}.

\begin{theorem} (Monge-Kantorovich) \cite{villani2008optimal} Let $\Omega^{(\mathcal{S})}$ and $\Omega^{(\mathcal{T})}$ be two separable metric spaces such that probability measures  $\boldsymbol{\mu}^{(\mathcal{S})}$ and $\boldsymbol{\mu}^{(\mathcal{T})}$ on $\Omega^{(\mathcal{S})}$ and $\Omega^{(\mathcal{T})}$, respectively, are Radon measures. The optimal coupling,
\begin{equation}
    P_* = \arginf_{P \in \Gamma(\boldsymbol{\mu}^{(\mathcal{S})},\boldsymbol{\mu}^{(\mathcal{T})})} \int_{\Omega^{(\mathcal{S})} \times \Omega^{(\mathcal{T})} } D(\boldsymbol{\mu}^{(\mathcal{S})},\boldsymbol{\mu}^{(\mathcal{T})}) \mathrm{d}P(\boldsymbol{\mu}^{(\mathcal{S})},\boldsymbol{\mu}^{(\mathcal{T})}) ,
\end{equation}
always exists for a distance function $D: \Omega^{(\mathcal{S})} \times \Omega^{(\mathcal{T})} \to [0,\infty)$, where $\Gamma$ is the set of all couplings (probability measures) on $\Omega^{(\mathcal{S})}$ and $\Omega^{(\mathcal{T})}$ with marginals $\boldsymbol{\mu}^{(\mathcal{S})}$ and $\boldsymbol{\mu}^{(\mathcal{T})}$, respectively. 
\label{thm:mk}
\end{theorem}

Intuitively, as illustrated in Figures~\ref{fig:ot} (a) and \ref{fig:costmap}, the OT problem attempts to determine the optimal way to move one probability distribution to another. If $\boldsymbol{\mu}^{(\mathcal{S})}$ and $\boldsymbol{\mu}^{(\mathcal{T})}$ constitute two datasets of size $N^{(\mathcal{S})}$ and $N^{(\mathcal{T})}$, respectively, there always exists an optimal probabilistic coupling $P_* \in \mathbb{R}^{N^{(\mathcal{S})} \times N^{(\mathcal{T})}}$ between the two datasets \cite{courty2017optimal}. Here, as shown in Figures~\ref{fig:costmap} where the source and target samples are assumed to separately follow bivariate distributions, $P_*$ is a doubly stochastic matrix---each row and column sums to one---that indicates the probability of a sample in the source match with all other points in the target. In occupancy mapping, $\boldsymbol{\mu}$ is computed as Dirac measures from LIDAR data (Appendix~I-A). 

With source data obtained in Section~\ref{sec:prep}, for a new target dataset, we attempt to obtain the optimal coupling, 
\begin{equation}
    P_* = \argmin_{P \in \Gamma(\mathbf{x}^{(\mathcal{S})},\mathbf{x}^{(\mathcal{T})})} \sum_{ij} P_{ij} D_{ij} - \lambda^{-1} r(P),
\label{eq:sink}
\end{equation}
for a given $D \in \mathbb{R}^{N^{(\mathcal{S})} \times N^{(\mathcal{T})}}$ distance matrix (e.g. squared Euclidean distance between source-target pairs) with the information entropy of $P$,
\begin{equation}
    r(P) = - \sum_{ij} P_{ij} \log P_{ij}.
\end{equation}

This entropic regularization, commonly known as the Sinkhorn distance \cite{cuturi2013sinkhorn,genevay2017learning}, enables solving the otherwise hard integer programming problem using an efficient iterative algorithm \cite{sinkhorn1967concerning}. Here, $\lambda$ controls the amount of regularization\footnote{$\lambda$ can be set to a large number depending on the machine precision of the computer.}. 

Having obtained the optimal coupling between source and target LIDAR, as illustrated in Figures~\ref{fig:ot} (b)-(c) and \ref{fig:algo}, now it is possible to transport source parameters $\theta^{(\mathcal{S})}$ to the target domain. This is done by associating the source parameter positions $\bar{\mathbf{h}}^{(\mathcal{S})}$ with source samples $\mathbf{x}^{(\mathcal{S})}$ as a linear map \cite{perrot2016mapping}, and transporting them to the target domain $\bar{\mathbf{h}}^{(\mathcal{S})} \to \bar{\mathbf{h}}^{(\mathcal{T})}$ according to the coupling matrix $P_*$ learned from LIDAR matching. All other $\theta^{(\mathcal{S})}$ parameters associated with the kernels positioned at $\bar{\mathbf{h}}^{(\mathcal{S})}$ will also be transported to the target domain. This implicit transfer process is depicted in Figure~\ref{fig:algo}.

\begin{figure}[t]
\begin{align*}
\text{learn } P_* \text{ for } & { \color{blue} \mathbf{x}^{(\mathcal{S})} } \xrightarrow[\text{transport}]{\text{explicit}} { \color{blue} \mathbf{x}^{(\mathcal{T})} } \\[5pt]
 \text{predict} {\color{white} --.}  & { \color{blue} \mathbf{\bar{h}}^{(\mathcal{S})} } \xrightarrow[\text{transport}]{\text{explicit}} { \color{red} \mathbf{\bar{h}}^{(\mathcal{T})} } \text{ using } P_* \\[-10pt]
  & \text{ } \vdots  {\color{white} ----...} \text{ } \vdots {\color{white} --} \\[-8pt]
{\color{white}   \text{-------}}   &  { \color{blue} \theta^{(\mathcal{S})} } \xrightarrow[\text{transport}]{\text{implicit}} { \color{red} \theta^{(\mathcal{T})} } \\
\end{align*}
\vspace{-30pt}
\caption{Parameter optimal transport. Known and unknown quantities are in blue and red, respectively. We learn an optimal coupling matrix $P_*$ using source and target LIDAR. Then we use this coupling matrix to predict target kernel positions corresponding to the source kernel position. By doing this, the other parameters associated with each kernel are also implicitly transported by treating them as labels.}
\label{fig:algo}
\end{figure}

\begin{figure*}[]
    \centering
    \includegraphics[width=0.9\textwidth]{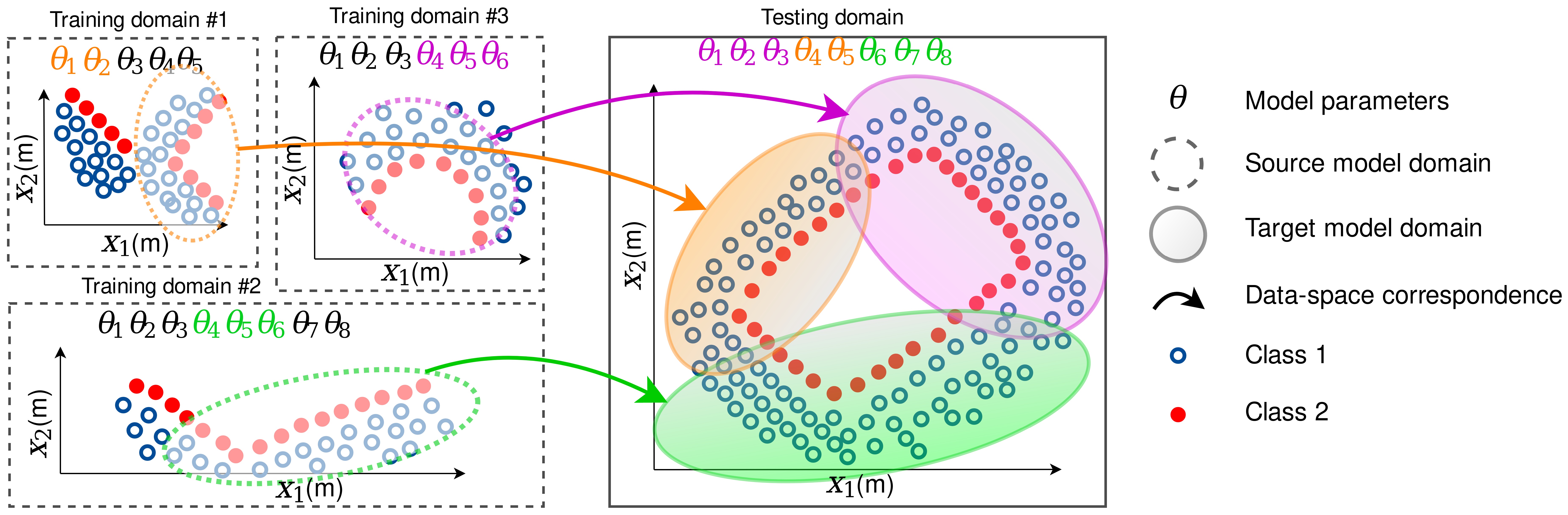}
    \caption{A high-level overview of the proposed method: Parameter Optimal Transport (POT). \textit{Training domains} correspond to potentially independent, data-intensive, expensive, yet small-scale pre-learned models. After storing in a dictionary of atoms, representative data-space and model-parameter tuples from the pre-learned set of models, we find data-space correspondences using optimal transport. These correspondences are then used to transport pre-learned parameters to out-of-sample \textit{test domains}.}
    \label{fig:method}
\end{figure*}




\subsection{Transport from a Dictionary of Atoms}
\label{sec:tda}

Although we created a dictionary of atoms consisting of diverse geometric primitives in Section~\ref{sec:prep}, the transfer procedure introduced in Section~\ref{sec:pot} was limited to a single LIDAR sector. In order to effectively make use of the entire dictionary, it is required to find the optimal coupling matrix over all elements in the dictionary $\mathbf{x}^{(\mathcal{S})} \in \mathcal{X}^{(\mathcal{S})}$. 

As another fact, although eq.~\ref{eq:sink} can be used to obtain a translation and scale invariant solution, it is not robust enough against large rotation variations. However, we can rotate data about the centroid of each atom using the rotation matrix, 
\begin{equation}
R(\alpha) = 
\begin{bmatrix}
\cos(\alpha) & -\sin(\alpha) \\
\sin(\alpha) & \quad \cos(\alpha)
\end{bmatrix},
\end{equation}
for a discrete set of rotations $\alpha \in \mathcal{A}$.


Overall, we obtain a candidate optimal coupling set of size $|\mathcal{X}^{(\mathcal{S})}| \times |\mathcal{A}|$ by minimizing eq.~\ref{eq:sink} over all rotations and atoms,
\begin{equation}
    \mathcal{P}_* = 
    \bigg\{
    \argmin_{
    P \in \Gamma(\mathbf{x}^{(\mathcal{S})},R(\alpha)\mathbf{x}^{(\mathcal{T})}) 
      } \sum_{ij} P_{ij} D_{ij} - \lambda^{-1} r(P) \bigg\}_{ \substack{
    \mathbf{x}^{(\mathcal{S})} \in \mathcal{X}^{(\mathcal{S})} \\
    \alpha \in \mathcal{A} \,
    } }.
\label{eq:sink_rot}
\end{equation}

Ultimately, we select the overall best coupling matrix from the candidate set $\mathcal{P}_*$ as the candidate that has the minimum 2-Wasserstein distance (refer Appendix~I-B) to the target, 
\begin{equation}
    P_* = \argmin_{P \in \mathcal{P}_*} {\sum_{ij} P_{ij} D_{ij}  }.
    \label{eq:sink_rot_opt}
\end{equation}

This $P_*$ can now be used to transfer parameters using the same method explained in Figure~\ref{fig:algo}. As a result of the computation procedure introduced in this section, as depicted in Figure~\ref{fig:method}, atoms from various domains will be transferred to the target. Because atoms only consist of a few hundred LIDAR points, this transfer can be performed in real-time. Unlike in BHM or ABHM, we can now introduce thousands of kernels. The increasing number of pre-learned kernels as well as the nonstationarity help to improve the accuracy. The entire Parameter Optimal Transport (POT) algorithm is summarized in Algorithm~\ref{algo:1}.


\subsection{POT Maps and Refined POT Maps}
\label{sec:repot}

Transporting parameters can be performed in two different ways. It is possible to transport parameters for each LIDAR scan separately, and immediately build the occupancy map. This results in an \emph{instantaneous map} which is useful for understanding the occupancy of the surrounding at present. Such maps can be used for safe decision-making and control in the locality of the robot. On the other hand, it is also possible to build the \emph{overall map} by sequentially aggregating the transported parameters as the robot moves. The overall map model completely discards training LIDAR data after transporting the parameters. This enables mapping large areas at a constant cost.

Once the parameters are transported with the intention of building an instantaneous or overall map, an occupancy map can be generated by plugging in the transported parameters to eq. (1) and querying occupancy probabilities. It will not only provide the mean occupancy map, but also the uncertainty as the variance estimate. Since only the parameters of the continuous mapping function eq.\ref{eq:bhm} are stored, the occupancy map can later be queried \emph{at any time at any resolution}.

Learning kernel parameters $\gamma$ and $\mathbf{h}$ in real-time is not feasible with ABHM. However, learning weights $w$, assuming other parameters are given, we have a fast approximation given by Bayesian Hilbert maps (BHMs) \cite{senanayake2017bayesian}. As an additional step to further improve the map quality, we propose to \textit{use transported parameters as prior distributions} of the BHM and simply update the weights $w$ by using \cite{senanayake2017bayesian}. We call this improved map, the refined POT (RePOT) map. 

\begin{algorithm}[t]
 \KwIn{New LIDAR scans, Source dictionary of atoms}
 \While{new scan in new domain}{
 $\mathcal{P}_* = \{ \}$\; 
  \For{each atom in $\mathcal{X}^{(\mathcal{S})}$}{ 
    \For{each rotation in $\mathcal{A}$}{
      $\mathcal{P}_*.\mathrm{insert(}$Compute the coupling matrix$\mathrm{)}$ (Eq.~\ref{eq:sink_rot})\;
    }
  }
  $P_* \leftarrow$ Determine the best coupling matrix (Eq.~\ref{eq:sink_rot_opt})\;
  $\theta^{(\mathcal{T})} \leftarrow $ Transfer the source parameters to the target domain using $P_*$ (Figure~\ref{fig:algo})\;
 }
 \KwOut{Parameters $\theta^{(\mathcal{T})}$}
 \caption{Transferring parameters to a new domain}
 \label{algo:1}
\end{algorithm}

\section{Experiments} 
\label{sec:experiments}

Both simulated and real-world datasets were used to assess the quality of POT. To generate simulated data, Carla v.0.9.2 simulator \cite{dosovitskiy2017carla} was used as it closely resembles real-world towns. As a real-world dataset, we used the KITTI benchmark dataset \cite{geiger2013vision}. All datasets are listed in Table~\ref{table:data} and each of these environments is considered as a domain. More details are provided in Appendix~II-A. As evaluation metrics, we used accuracy (ACC), area under ROC curve (AUC) and negative log-likelihood (NLL) \cite{Bishop2006}. Unlike ACC and AUC, NLL takes into account uncertainty of predictions. The higher the AUC or lower the NLL, the better the model is. 

\begin{table}[t]
    \centering
    \caption{Description of domains}
        \begin{tabular}{l|l}
        \toprule
            \multicolumn{1}{c|}{Domains (Datasets)}  & \multicolumn{1}{c}{Description} \\
            \hline
            Carla Town 1 & a 2D dataset in town 1 in Carla (3.7 km).\\
            Carla Town 2 & a 2D dataset in town 1 in Carla (1.5 km).\\
            Carla Town 3 & a 2D dataset in town 1 in Carla (8.6 km).\\
            Carla Town 1 3D & a 3D dataset in town 1 in Carla.\\
            Carla Town 1 Dyna & Carla Town 1 with 120 vehicles running around. \\
            KITTI Dyna & a 2D dataset (the middle LIDAR channel). \\
        \bottomrule
        \end{tabular}
    \label{table:data}
\end{table}
\vspace{-2pt}
 
\subsection{Intra-domain and Inter-domain Adaptation}

\begin{figure}[]
    \centering
    \includegraphics[width=1.0\linewidth]{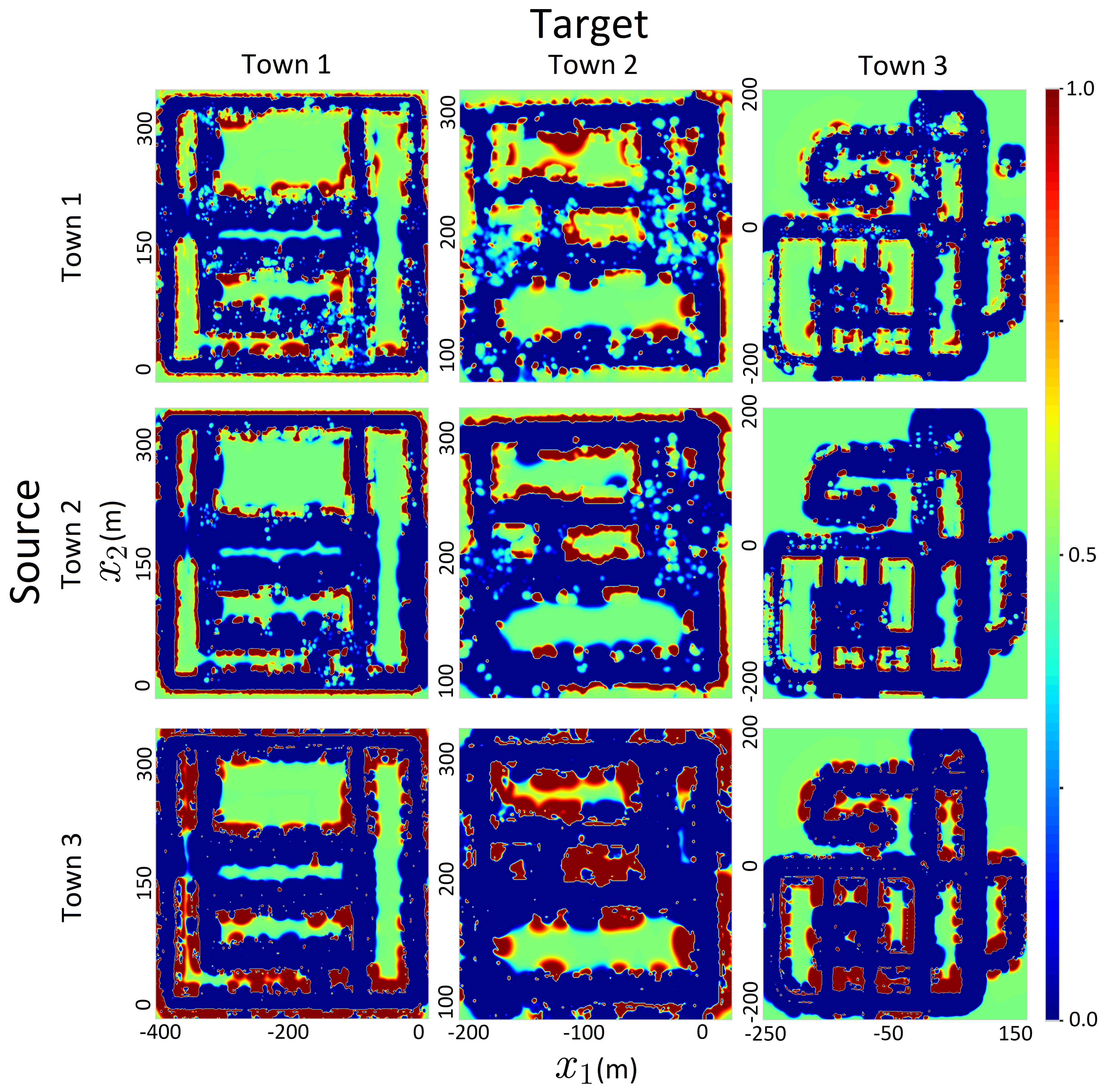}
    \caption{Transported occupancy maps for the inter and intra domain adaptation experiments using the town datasets. From top to bottom and left to right are towns 1, 2, and 3.}
    \label{fig:crossdomain_matrix}
\end{figure}

\begin{figure*}[]
    \centering
    \includegraphics[width=0.9\linewidth]{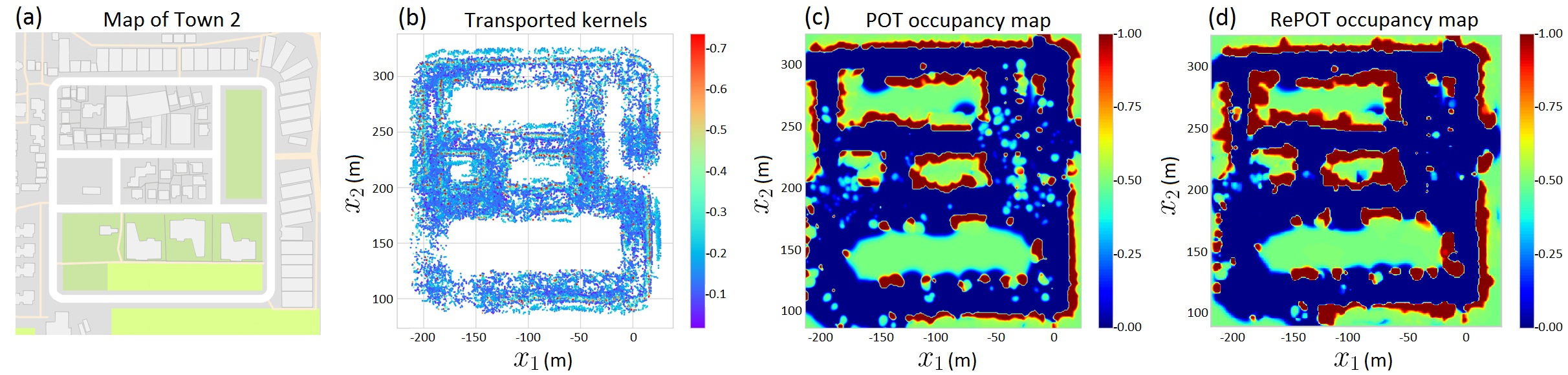}
    \caption{Large-scale map building with POT and RePOT. (a) Carla Town 2 plan. (b) Transferred kernel mean width and position parameters. (c) Occupancy prediction with POT. (d) Occupancy prediction with RePOT.}
    \label{fig:carla_town1_full}
\end{figure*}

In this experiment, we consider two paradigms: intra-domain and inter-domain transfer. In intra-domain transfer, the source atoms are generated from the first 10 frames of a particular dataset and parameters are transferred to the rest of the same dataset while they are transferred to a completely different domain in inter-domain transfer. Based on results reported in Table~\ref{table:comp} with 20\% randomly sampled test LIDAR beams from each town, it is possible to accurately transfer parameters using POT. This enables mapping large scale towns in real-time. All parameters are aggregated over time to build occupancy maps of the entire environments as visualized in Figure~\ref{fig:crossdomain_matrix} and Appendix~II-C. Using the Town 1 3D dataset, we demonstrate the possibility of extending POT to 3D environments. In this case, source atoms described in Section~\ref{sec:prep}, were circular cylindrical sectors (i.e. pie slice shaped). The post-hoc refinement procedure, RePOT, introduced in Section~\ref{sec:repot}, further improved the map significantly. A visualization of RePOT is shown in Figure~\ref{fig:carla_town1_full} and performance improvement, in direct comparison with results in Table~\ref{table:comp}, is reported in Table~\ref{table:comp_repot}.
\vspace{-1pt}

\begin{table*}[]
\centering
\makebox[0pt][c]{\parbox{1.0\textwidth}{%
    \begin{minipage}[b]{0.55\hsize}\centering
        \caption{Performance of intra-domain (diagonal entries of the table) and inter-domain (off-diagonal entries of the table) transfer.}
    \begin{tabular}{cc||c|c|c|c}
    \toprule
     &&&\multicolumn{3}{c}{\bf Target}  \\
       & & & Town1 & Town2 & Town3\\
      \cline{2-6}
            \parbox[t]{2mm}{\multirow{9}{*}{\rotatebox[origin=c]{90}{\bf Source}}}
      &\parbox[t]{2mm}{\multirow{3}{*}{\rotatebox[origin=c]{90}{ACC}}}  &Town1 & 0.79  & 0.82 & 0.76 \\
     & &Town2 & 0.70 & 0.72 & 0.58  \\
      &&Town3 & 0.85 & 0.83 & 0.84  \\
      \cline{2-6}
     & \parbox[t]{2mm}{\multirow{3}{*}{\rotatebox[origin=c]{90}{AUC}}}  &Town1 & 0.88 & 0.88 & 0.90 \\
      &&Town2 & 0.85 & 0.83 & 0.83  \\
      &&Town3 & 0.92 & 0.92 & 0.93  \\
      \cline{2-6}
      &\parbox[t]{2mm}{\multirow{3}{*}{\rotatebox[origin=c]{90}{NLL}}} &Town1 & 1.14 & 0.97 & 1.40\\
      &&Town2 & 3.30 & 3.23 & 5.98  \\
     & &Town3 & 1.64 & 1.69 & 1.79  \\
      \bottomrule
    \end{tabular}
  \label{table:comp}
    \end{minipage}
    \hfill
    \begin{minipage}[b]{0.45\hsize}\centering
    \caption{Performance metrics of Refined POT (RePOT) across both intra- and inter-domain transfers.}
    \begin{tabular}{cc||c|c|c|c}
    \toprule
     &&&\multicolumn{3}{c}{\bf Target}  \\
       & & & Town1 & Town2 & Town3\\
      \cline{2-6}
            \parbox[t]{2mm}{\multirow{9}{*}{\rotatebox[origin=c]{90}{\bf Source}}}
      &\parbox[t]{2mm}{\multirow{3}{*}{\rotatebox[origin=c]{90}{ACC}}}  &Town1 & 0.95 & 0.93 & 0.95 \\
     & &Town2 & 0.91 & 0.91 & 0.92  \\
      &&Town3 & 0.95 & 0.92 & 0.93  \\
      \cline{2-6}
     & \parbox[t]{2mm}{\multirow{3}{*}{\rotatebox[origin=c]{90}{AUC}}}  &Town1 & 0.99 & 0.98 & 0.98 \\
      &&Town2 & 0.98 & 0.98 & 0.98  \\
      &&Town3 & 0.99 & 0.97 & 0.97  \\
      \cline{2-6}
      &\parbox[t]{2mm}{\multirow{3}{*}{\rotatebox[origin=c]{90}{NLL}}} &Town1 & 0.71 & 1.4 & 1.12 \\
      &&Town2 & 1.40 & 1.74 & 1.85  \\
     & &Town3 & 0.96 & 1.62 & 1.44  \\
      \bottomrule
    \end{tabular}
  \label{table:comp_repot}
    \end{minipage}
}}
\end{table*}

\subsection{Building Instantaneous Maps}
\label{sec:ins}
This experiment was designed to demonstrate how parameters can be instantaneously transported to build the instantaneous map of the surrounding. For this purpose, we used the two dynamic environments: Town 1 Dyna and KITTI Dyna. The source dictionary of atoms was prepared similar to the intra/inter-domain adaptation experiment. Such a map is shown in Figure~\ref{fig:kitti_frame}. The performance of the model was evaluated on 20\% of data that were not used for optimal transport. Table~\ref{table:ins} shows the performance of transferring features extracted from each town to the dynamic datasets. 

\begin{table}[]
\centering
    \caption{Instantaneous maps in dynamic environments: Experiments for sim2sim and sim2real with mean and SD.}
    \begin{tabular}{cc||c|c|c} 
    \toprule
     &&&\multicolumn{2}{c}{\bf Target}  \\
     &&& Town 1 Dyna & KITTI Dyna  \\
     \cline{2-5}
     \parbox[t]{2mm}{\multirow{9}{*}{\rotatebox[origin=c]{90}{\bf Source}}}
    &\parbox[t]{2mm}{\multirow{3}{*}{\rotatebox[origin=c]{90}{ACC}}}
      &Town 1  & 0.74 $\pm$ 0.10 & 0.69 $\pm$ 0.06 \\
      &&Town 2  & 0.70 $\pm$ 0.10 & 0.58 $\pm$ 0.06\\
      &&Town 3  & 0.74 $\pm$ 0.11 & 0.71 $\pm$ 0.07\\
      \cline{2-5}
      &\parbox[t]{2mm}{\multirow{3}{*}{\rotatebox[origin=c]{90}{AUC}}}
     &Town 1  & 0.81 $\pm$0.11 & 0.77 $\pm$0.06\\
     &&Town 2  & 0.77 $\pm$0.12 & 0.73 $\pm$0.06\\
     &&Town 3  & 0.78 $\pm$0.15 & 0.73 $\pm$0.09\\
     \cline{2-5}
     &\parbox[t]{2mm}{\multirow{3}{*}{\rotatebox[origin=c]{90}{NLL}}}
     &Town 1  & 1.06 $\pm$ 0.56 & 1.42 $\pm$ 0.38 \\
     &&Town 2  & 1.90 $\pm$ 0.79 & 3.63 $\pm$ 1.04 \\
     &&Town 3  & 1.89 $\pm$ 1.30 & 2.30 $\pm$ 0.83\\
      \bottomrule
    \end{tabular}
  \label{table:ins}
  \vspace{-5pt}
\end{table}

\subsection{Performance Comparison}

\begin{table}[]
\centering
   \caption{Performance per time unit for RePOT, POT, ABHM, and BHM. Though OGM results are reported for reference purposes, unlike other methods, OGM cannot be computed for per time unit basis.}
    \begin{tabular}{c||c|c|c|c}
    \toprule
     &&\multicolumn{3}{c}{\bf Target}  \\
       & Method & Town1 & Town2 & Town3\\
      \hline
      \parbox[t]{2mm}{\multirow{5}{*}{\rotatebox[origin=c]{90}{ACC}}} 
      &RePOT  & 0.95 & 0.93 & 0.95 \\
      &POT  & 0.85 & 0.83 & 0.84 \\  
      &ABHM & 0.77 & 0.59 & 0.86  \\
      &BHM & 0.66 & 0.61 & 0.71 \\
      &OGM & 0.78 & 0.78 & 0.77 \\
      \hline
      \parbox[t]{2mm}{\multirow{5}{*}{\rotatebox[origin=c]{90}{AUC}}} 
     &RePOT  & 0.99 & 0.98 & 0.98 \\
     &POT & 0.92 & 0.92 & 0.93 \\
     &ABHM & 0.95 & 0.96 & 0.96  \\
      &BHM & 0.94 & 0.92 & 0.91 \\
      &OGM & 0.89 & 0.91 & 0.90 \\
      \hline
      \parbox[t]{2mm}{\multirow{5}{*}{\rotatebox[origin=c]{90}{NLL}}}
    &RePOT  & 0.71 & 1.41 & 1.12 \\
    &POT &1.64& 1.69 & 1.79 \\
      &ABHM & 0.58 & 0.71 & 0.41  \\
     &BHM & 0.63 & 0.69 & 0.61 \\
     &OGM & 2.00 & 1.34 & 1.13 \\
      \bottomrule
    \end{tabular}
    \label{table:compwith}
    \vspace*{-0.4cm}
\end{table}

In this experiment, we compared various occupancy mapping algorithms in terms of accuracy and speed. Since these algorithms cannot be trained or queried in a similar fashion, we measured the per time unit performance. For instance, ABHM can only be trained in small environments although our datasets consist of large towns. Firstly, we measure the time for running POT per LIDAR scan. Then we decide the number of kernels to match the same runtime for BHM and ABHM. Results are reported in Table~\ref{table:compwith}. Though OGM cannot be computed per time basis, we report the results for reference (See Appendix~II-B). GPOM cannot be executed for datasets this large. As expected, ABHM outperforms BHM in all metrics because ABHM is a nonstationary model that takes into account local geometry. Theoretically, in the infinite memory and computation time limit, ABHM should outperform all methods. Nonetheless, practically, POT has a higher ACC and AUC compared to ABHM as POT can transfer kernels online to accommodate the complexity of the environment. However, the increase in NLL in POT compared to ABHM, indicates the inherent uncertainties of the transfer procedure. Once the weights were refined using RePOT, NLL has dropped as the weight distributions can be optimized to reduce the uncertainty giving better predictions.

{\bf Runtime:} With a laptop with 4 cores and 8 GB RAM, on average, POT, programmed in Python, takes around 1 s update time. This is without parallelizing any part of the code. Note that eq. (\ref{eq:sink_rot}) is highly parallelizable making the algorithm $|\mathcal{X}^{(\mathcal{S})}| \times |\mathcal{A}|$ faster (approx. 25 times). This is a significant improvement to algorithms such as BHM and ABHM which would take several hours to build a large-scale map as they rely on complicated variational inference procedures. POT run-time increases with increasing $\lambda$ in the Sinkhorn algorithm we used in POT. As $\lambda \to \infty$ the convergence is guaranteed.

\section{Discussion} 
\label{sec:discussion}
In optimal transport, we consider the problem of transforming one probability measure to another. This also loosely relates to the point cloud registration problem typically addressed by the iterative closest point (ICP) algorithm \cite{besl1992method}. However, unlike ICP which only has a single set of translation and rotation parameters, in optimal transport, each data point in the source dataset has a highly nonlinear relationship with every other point in target datapoints through the optimal coupling matrix $P_*$. Another reason why we cannot resort to a popular algorithm such as ICP is because it only works for slight changes in translation and rotation. When a robot moves in dynamic environments, it is essential to adapt for sudden, potentially large, nonlinear changes in geometry.

One remarkable aspect of being able to transport distributions is that it endows us the ability to adapt Bayesian models in the sense of an \textit{informed prior} \cite{gronau2017bayesian} enabling expedited parameter tuning. We have demonstrated such a use case in RePOT with significant improvements in overall map quality.

Although our method was presented and demonstrated in the context of occupancy mapping, there are many other potential applications in robotics. For example, the theory can be potentially used for domain adaptation of policy parameters where a policy is trained in one environment and needs to be transferred to another. For example, a particular robotic arm is trained to grasp objects on a table and performs well on this task. One could, in principle adapt policies for use in another arm without retraining the policy from the start. Finally, it can also be used for sim2real where models are learned in simulation and transferred to the physical world, saving significant time and cost in running real robots.  

\section{Conclusion} 
\label{sec:conclusion}
This paper introduced parameter optimal transport (POT), an efficient framework for geometric domain adaptation. By combining the formalism of automorphing Bayesian Hilbert maps with optimal transport theory, patterns from one environment can be seamlessly transferred to another in a fraction of a second. We show that this framework can be effectively used to map large urban environments, transferring learned patterns between two cities, between simulated and real environments, and between static and dynamic environments.


\bibliographystyle{IEEEtran.bst}
\bibliography{references}

\begin{thebibliography}{10}
\providecommand{\url}[1]{#1}
\csname url@samestyle\endcsname
\providecommand{\newblock}{\relax}
\providecommand{\bibinfo}[2]{#2}
\providecommand{\BIBentrySTDinterwordspacing}{\spaceskip=0pt\relax}
\providecommand{\BIBentryALTinterwordstretchfactor}{4}
\providecommand{\BIBentryALTinterwordspacing}{\spaceskip=\fontdimen2\font plus
\BIBentryALTinterwordstretchfactor\fontdimen3\font minus
  \fontdimen4\font\relax}
\providecommand{\BIBforeignlanguage}[2]{{%
\expandafter\ifx\csname l@#1\endcsname\relax
\typeout{** WARNING: IEEEtran.bst: No hyphenation pattern has been}%
\typeout{** loaded for the language `#1'. Using the pattern for}%
\typeout{** the default language instead.}%
\else
\language=\csname l@#1\endcsname
\fi
#2}}
\providecommand{\BIBdecl}{\relax}
\BIBdecl

\bibitem{ammar2015autonomous}
H.~B. Ammar, E.~Eaton, J.~M. Luna, and P.~Ruvolo, ``Autonomous cross-domain
  knowledge transfer in lifelong policy gradient reinforcement learning,'' in
  \emph{International Joint Conference on Artificial Intelligence (IJCAI)},
  2015, pp. 3345--3351.

\bibitem{sadeghi2017sim2real}
------, ``Autonomous cross-domain knowledge transfer in lifelong policy
  gradient reinforcement learning,'' in \emph{IEEE Computer Society Conference
  on Computer Vision and Pattern Recognition (CVPR)}, 2018.

\bibitem{finn2017model}
C.~Finn, P.~Abbeel, and S.~Levine, ``Model-agnostic meta-learning for fast
  adaptation of deep networks,'' in \emph{International Conference on Machine
  Learning (ICML)}, 2017, pp. 1126--1135.

\bibitem{fang2018dart}
X.~Fang, H.~Bai, Z.~Guo, B.~Shen, S.~Hoi, and Z.~Xu, ``Dart: Domain-adversarial
  residual-transfer networks for unsupervised cross-domain image
  classification,'' \emph{arXiv preprint arXiv:1812.11478}, 2018.

\bibitem{wulfmeier2018incremental}
M.~Wulfmeier, A.~Bewley, and I.~Posner, ``Incremental adversarial domain
  adaptation for continually changing environments,'' in \emph{IEEE
  International Conference on Robotics and Automation (ICRA)}, 2018, pp. 1--9.

\bibitem{meier2014efficient}
F.~Meier, P.~Hennig, and S.~Schaal, ``Efficient bayesian local model learning
  for control,'' in \emph{IEEE/RSJ International Conference on Intelligent
  Robots and Systems (IROS)}.\hskip 1em plus 0.5em minus 0.4em\relax IEEE,
  2014, pp. 2244--2249.

\bibitem{deisenroth2011pilco}
M.~Deisenroth and C.~E. Rasmussen, ``Pilco: A model-based and data-efficient
  approach to policy search,'' in \emph{International Conference on Machine
  Learning (ICML)}, 2011, pp. 465--472.

\bibitem{wuthrich-rgf-2015}
M.~W{\"u}thrich, C.~Garcia~Cifuentes, S.~Trimpe, F.~Meier, J.~Bohg, J.~Issac,
  and S.~Schaal, ``Robust gaussian filtering using a pseudo measurement,'' in
  \emph{American Control Conference (ACC)}, 2016.

\bibitem{campbell2017bayesian}
J.~Campbell and H.~B. Amor, ``Bayesian interaction primitives: A slam approach
  to human-robot interaction,'' in \emph{Conference on Robot Learning (CoRL)},
  2017, pp. 379--387.

\bibitem{burchfiel2017bayesian}
B.~Burchfiel and G.~Konidaris, ``Bayesian eigenobjects: A unified framework for
  3d robot perception.'' in \emph{Robotics: Science and Systems (RSS)}, 2017.

\bibitem{unhelkar2018learning}
V.~V. Unhelkar and J.~A. Shah, ``Learning models of sequential decision-making
  without complete state specification using bayesian nonparametric inference
  and active querying,'' \emph{Massachusetts Institute of Technology}, 2018.

\bibitem{isele2016using}
D.~Isele, M.~Rostami, and E.~Eaton, ``Using task features for zero-shot
  knowledge transfer in lifelong learning.'' in \emph{International Joint
  Conference on Artificial Intelligence (IJCAI)}, 2016, pp. 1620--1626.

\bibitem{tomkins2018nonstationar}
R.~Senanayake, A.~Tompkins, and F.~Ramos, ``Automorphing kernels for
  nonstationarity in mapping unstructured environments,'' in \emph{Conference
  on Robot Learning (CoRL)}, 2018, pp. 443--455.

\bibitem{lasota2017survey}
P.~A. Lasota, T.~Fong, J.~A. Shah \emph{et~al.}, ``A survey of methods for safe
  human-robot interaction,'' \emph{Foundations and Trends in Robotics}, vol.~5,
  no.~4, pp. 261--349, 2017.

\bibitem{akametalu2014reachability}
A.~K. Akametalu, S.~Kaynama, J.~F. Fisac, M.~N. Zeilinger, J.~H. Gillula, and
  C.~J. Tomlin, ``Reachability-based safe learning with gaussian processes,''
  in \emph{IEEE Conference on Decision and Control (CDC)}, 2014, pp. 443--455.

\bibitem{vallicrosa2018h}
G.~Vallicrosa and P.~Ridao, ``H-slam: Rao-blackwellized particle filter slam
  using hilbert maps,'' \emph{Sensors}, vol.~18, no.~5, 2018.

\bibitem{villani2008optimal}
C.~Villani, \emph{Optimal transport: old and new}.\hskip 1em plus 0.5em minus
  0.4em\relax Springer Science \& Business Media, 2008, vol. 338.

\bibitem{arjovsky2017wasserstein}
M.~Arjovsky, S.~Chintala, and L.~Bottou, ``Wasserstein gan,'' \emph{arXiv
  preprint arXiv:1701.07875}, 2017.

\bibitem{solomon2015convolutional}
J.~Solomon, F.~De~Goes, G.~Peyr{\'e}, M.~Cuturi, A.~Butscher, A.~Nguyen, T.~Du,
  and L.~Guibas, ``Convolutional wasserstein distances: Efficient optimal
  transportation on geometric domains,'' \emph{ACM Transactions on Graphics
  (TOG)}, vol.~34, no.~4, pp. 1--11, 2015.

\bibitem{GPOMIJRR}
S.~T. O'Callaghan and F.~T. Ramos, ``Gaussian process occupancy maps,''
  \emph{International Journal of Robotics Research (IJRR)}, vol.~31, no.~1, pp.
  42--62, 2012.

\bibitem{wang2016fast}
J.~Wang and B.~Englot, ``Fast, accurate gaussian process occupancy maps via
  test-data octrees and nested bayesian fusion,'' in \emph{IEEE International
  Conference on Robotics and Automation (ICRA)}, 2016, pp. 1003--1010.

\bibitem{ElfesThesis}
A.~Elfes, ``Occupancy grids: a probabilistic framework for robot perception and
  navigation,'' Ph.D. dissertation, Carnegie Mellon University, 1989.

\bibitem{arbuckle2002temporal}
D.~Arbuckle, A.~Howard, and M.~Mataric, ``Temporal occupancy grids: a method
  for classifying the spatio-temporal properties of the environment,'' in
  \emph{IEEE/RSJ International Conference on Intelligent Robots and Systems
  (IROS)}, vol.~1, 2002, pp. 409--414.

\bibitem{Ramos15}
F.~Ramos and L.~Ott, ``Hilbert maps: scalable continuous occupancy mapping with
  stochastic gradient descent,'' in \emph{Robotics: Science and Systems (RSS)},
  2015.

\bibitem{rans2017}
R.~Senanayake, S.~O'Callaghan, and F.~Ramos, ``Learning highly dynamic
  environments with stochastic variational inference,'' in \emph{IEEE
  International Conference on Robotics and Automation (ICRA)}, 2017.

\bibitem{senanayake2016spatio}
R.~Senanayake, L.~Ott, S.~O'Callaghan, and F.~T. Ramos, ``Spatio-temporal
  hilbert maps for continuous occupancy representation in dynamic
  environments,'' in \emph{Advances in Neural Information Processing Systems
  (NIPS)}, 2016, pp. 3925--3933.

\bibitem{senanayake2017bayesian}
R.~Senanayake and F.~Ramos, ``Bayesian hilbert maps for dynamic continuous
  occupancy mapping,'' in \emph{Conference on Robot Learning (CoRL)}, 2017, pp.
  458--471.

\bibitem{hofmann2008kernel}
T.~Hofmann, B.~Sch{\"o}lkopf, and A.~J. Smola, ``Kernel methods in machine
  learning,'' \emph{The Annals of Statistics}, pp. 1171--1220, 2008.

\bibitem{senanayake2018building}
R.~Senanayake and F.~Ramos, ``Building continuous occupancy maps with moving
  robots,'' in \emph{AAAI Conference on Artificial Intelligence (AAAI)}, 2018.

\bibitem{pan2009survey}
S.~J. Pan and Q.~Yang, ``A survey on transfer learning,'' \emph{IEEE
  Transactions on Knowledge and Data Engineering}, vol.~22, no.~10, pp.
  1345--1359, 2009.

\bibitem{bousmalis2018using}
K.~Bousmalis, A.~Irpan, P.~Wohlhart, Y.~Bai, M.~Kelcey, M.~Kalakrishnan,
  L.~Downs, J.~Ibarz, P.~Pastor, K.~Konolige \emph{et~al.}, ``Using simulation
  and domain adaptation to improve efficiency of deep robotic grasping,'' in
  \emph{IEEE International Conference on Robotics and Automation (ICRA)}.\hskip
  1em plus 0.5em minus 0.4em\relax IEEE, 2018, pp. 4243--4250.

\bibitem{hoffman2016fcns}
J.~Hoffman, D.~Wang, F.~Yu, and T.~Darrell, ``Fcns in the wild: Pixel-level
  adversarial and constraint-based adaptation,'' \emph{arXiv preprint
  arXiv:1612.02649}, 2016.

\bibitem{ghifary2016deep}
M.~Ghifary, W.~B. Kleijn, M.~Zhang, D.~Balduzzi, and W.~Li, ``Deep
  reconstruction-classification networks for unsupervised domain adaptation,''
  in \emph{European Conference on Computer Vision (ECCV)}.\hskip 1em plus 0.5em
  minus 0.4em\relax Springer, 2016, pp. 597--613.

\bibitem{zhu2017unpaired}
J.-Y. Zhu, T.~Park, P.~Isola, and A.~A. Efros, ``Unpaired image-to-image
  translation using cycle-consistent adversarial networks,''
  \emph{International Conference on Computer Vision (ICCV)}, 2017.

\bibitem{kim2017learning}
T.~Kim, M.~Cha, H.~Kim, J.~K. Lee, and J.~Kim, ``Learning to discover
  cross-domain relations with generative adversarial networks,'' in
  \emph{International Conference on Machine Learning (ICML)}, 2017, pp.
  1857--1865.

\bibitem{liu2017unsupervised}
M.-Y. Liu, T.~Breuel, and J.~Kautz, ``Unsupervised image-to-image translation
  networks,'' in \emph{Advances in Neural Information Processing Systems
  (NIPS)}, 2017, pp. 700--708.

\bibitem{courty2017optimal}
N.~Courty, R.~Flamary, D.~Tuia, and A.~Rakotomamonjy, ``Optimal transport for
  domain adaptation,'' \emph{IEEE Transactions on Pattern Analysis and Machine
  Intelligence (PAMI)}, vol.~39, no.~9, pp. 1853--1865, 2017.

\bibitem{cuturi2013sinkhorn}
M.~Cuturi, ``Sinkhorn distances: Lightspeed computation of optimal transport,''
  in \emph{Advances in Neural Information Processing Systems (NIPS)}, 2013, pp.
  2292--2300.

\bibitem{genevay2017learning}
A.~Genevay, G.~Peyr{\'e}, and M.~Cuturi, ``Learning generative models with
  sinkhorn divergences,'' \emph{1608-1617}, 2018.

\bibitem{sinkhorn1967concerning}
R.~Sinkhorn and P.~Knopp, ``Concerning nonnegative matrices and doubly
  stochastic matrices,'' \emph{Pacific Journal of Mathematics}, vol.~21, no.~2,
  pp. 343--348, 1967.

\bibitem{perrot2016mapping}
M.~Perrot, N.~Courty, R.~Flamary, and A.~Habrard, ``Mapping estimation for
  discrete optimal transport,'' in \emph{Advances in Neural Information
  Processing Systems (NIPS)}, 2016, pp. 4197--4205.

\bibitem{dosovitskiy2017carla}
A.~Dosovitskiy, G.~Ros, F.~Codevilla, A.~Lopez, and V.~Koltun, ``Carla: An open
  urban driving simulator,'' \emph{arXiv preprint arXiv:1711.03938}, 2017.

\bibitem{geiger2013vision}
A.~Geiger, P.~Lenz, C.~Stiller, and R.~Urtasun, ``Vision meets robotics: The
  kitti dataset,'' \emph{International Journal of Robotics Research (IJRR)},
  vol.~32, no.~11, pp. 1231--1237, 2013.

\bibitem{Bishop2006}
C.~Bishop, \emph{Pattern Recognition and Machine Learning}.\hskip 1em plus
  0.5em minus 0.4em\relax Springer, 2006.

\bibitem{besl1992method}
P.~J. Besl and N.~D. McKay, ``Method for registration of 3-d shapes,'' in
  \emph{Sensor fusion IV: control paradigms and data structures}, vol.
  1611.\hskip 1em plus 0.5em minus 0.4em\relax International Society for Optics
  and Photonics, 1992, pp. 586--606.

\bibitem{gronau2017bayesian}
Q.~F. Gronau, S.~Van~Erp, D.~W. Heck, J.~Cesario, K.~J. Jonas, and E.-J.
  Wagenmakers, ``A bayesian model-averaged meta-analysis of the power pose
  effect with informed and default priors: The case of felt power,''
  \emph{Comprehensive Results in Social Psychology}, vol.~2, no.~1, pp.
  123--138, 2017.

\end{thebibliography}

\end{document}